\pdfoutput=1

\documentclass[11pt]{article}

\usepackage[]{EMNLP2023}

\usepackage{times}
\usepackage{subcaption}
\usepackage{graphicx}
\usepackage{latexsym}
\usepackage{booktabs}
\usepackage{siunitx} 
\usepackage{fontsize}
\usepackage{caption}
\usepackage{hyperref}

\usepackage[T1]{fontenc}

\usepackage[utf8]{inputenc}

\usepackage{microtype}

\usepackage{inconsolata}

\title{GPT4All: An Ecosystem of Open Source Compressed Language Models}

  \author{Yuvanesh Anand\\
  Nomic AI\\
  \texttt{yuvanesh@nomic.ai} \\\And
  Zach Nussbaum \\
  Nomic AI\\
  \texttt{zach@nomic.ai} \\\And
  Adam Treat\\
  Nomic AI\\
  \texttt{adam@nomic.ai} \\\And
  Aaron Miller\\
  Nomic AI\\
  \texttt{aaron@nomic.ai} \\\AND
  Richard Guo\\
  Nomic AI\\
  \texttt{richard@nomic.ai} \\\And
  Ben Schmidt\\
  Nomic AI\\
  \texttt{ben@nomic.ai} \\\And
  GPT4All Community\\
  Planet Earth\\\AND
  Brandon Duderstadt\thanks{\hspace{5pt}Shared Senior Authorship}\\
  Nomic AI\\
  \texttt{brandon@nomic.ai} \\\And
  Andriy Mulyar\footnotemark[1]\\
  Nomic AI\\
  \texttt{andriy@nomic.ai}}

\begin{document}

\maketitle

\begin{abstract}
Large language models (LLMs) have recently achieved human-level performance on a range of professional and academic benchmarks.
The accessibility of these models has lagged behind their performance.
State-of-the-art LLMs require costly infrastructure; are only accessible via rate-limited, geo-locked, and censored web interfaces; and lack publicly available code and technical reports.

In this paper, we tell the story of GPT4All, a popular open source repository that aims to democratize access to LLMs.
We outline the technical details of the original GPT4All model family, as well as the evolution of the GPT4All project from a single model into a fully fledged open source ecosystem.
It is our hope that this paper acts as both a technical overview of the original GPT4All models as well as a case study on the subsequent growth of the GPT4All open source ecosystem. 
\end{abstract}

\section{Introduction}
On March 14 2023, OpenAI released GPT-4, a large language model capable of achieving human level performance on a variety of professional and academic benchmarks.
Despite the popularity of the release, the GPT-4 technical report~\citep{openai2023gpt4} contained virtually no details regarding the architecture, hardware, training compute, dataset construction, or training method used to create the model.
Moreover, users could only access the model through the internet interface at chat.openai.com, which was severely rate limited and unavailable in several locales (e.g. Italy)~\cite{bbc2023chatgpt}.
Additionally, GPT-4 refused to answer a wide variety of queries, responding only with the now infamous "As an AI Language Model, I cannot..." prefix \cite{verge2023ai}.
These transparency and accessibility concerns spurred several developers to begin creating open source large language model (LLM) alternatives.
Several grassroots efforts focused on fine tuning Meta's open code LLaMA model~\cite{touvron2023llama, wsj_llama}, whose weights were leaked on BitTorrent less than a week prior to the release of GPT-4~\cite{verge-meta-ai-leak-2023}.
GPT4All started as one of these variants.

In this paper, we tell the story of GPT4All. 
We comment on the technical details of the original GPT4All model~\cite{gpt4all}, as well as the evolution of GPT4All from a single model to an ecosystem of several models.
We remark on the impact that the project has had on the open source community, and discuss future directions.
It is our hope that this paper acts as both a technical overview of the original GPT4All models as well as a case study on the subsequent growth of the GPT4All open source ecosystem.

\section{The Original GPT4All Model}

\subsection{Data Collection and Curation}
To train the original GPT4All model, we collected roughly one million prompt-response pairs using the GPT-3.5-Turbo OpenAI API between March 20, 2023 and March 26th, 2023.
In particular, we gathered GPT-3.5-Turbo responses to prompts of three publicly available datasets: the unified chip2 subset of LAION OIG, a random sub-sample of Stackoverflow Questions, and a sub-sample of Bigscience/P3~\cite{sanh2021multitask}. Following the approach in Stanford Alpaca~\cite{alpaca}, an open source LLaMA variant that came just before GPT4All, we focused substantial effort on dataset curation.

\begin{figure*}[t]
  \begin{subfigure}[b]{.5\columnwidth}
        \centering
        \href{https://atlas.nomic.ai/map/gpt4all_data_clean}{\includegraphics[width=1\textwidth]{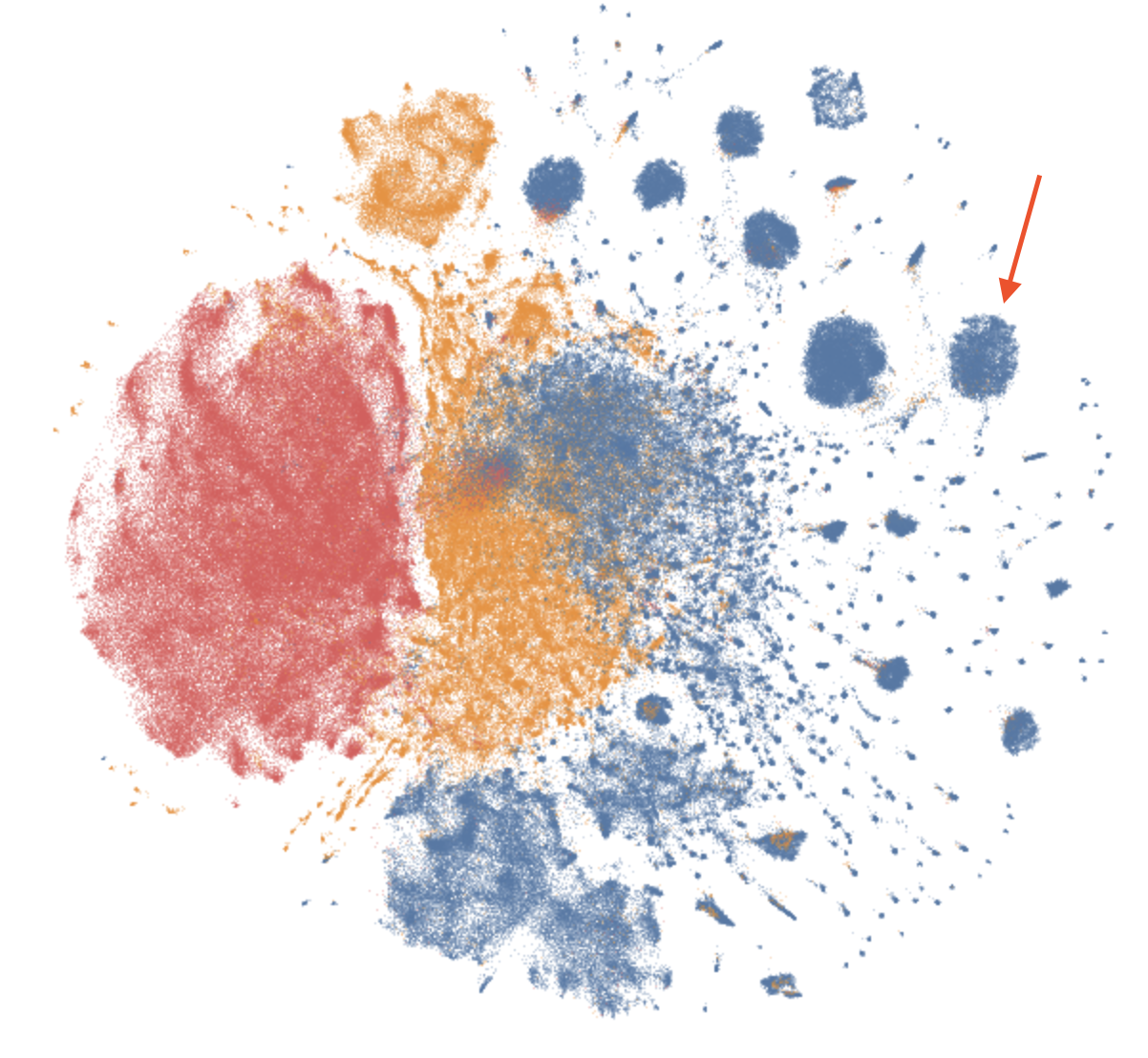}}
        \caption{}
        \label{subfig:dirty}
    \end{subfigure}
    \begin{subfigure}[b]{.45\columnwidth}
        \centering
        \href{https://atlas.nomic.ai/map/gpt4all_data_clean_without_p3}{\includegraphics[width=1\textwidth]{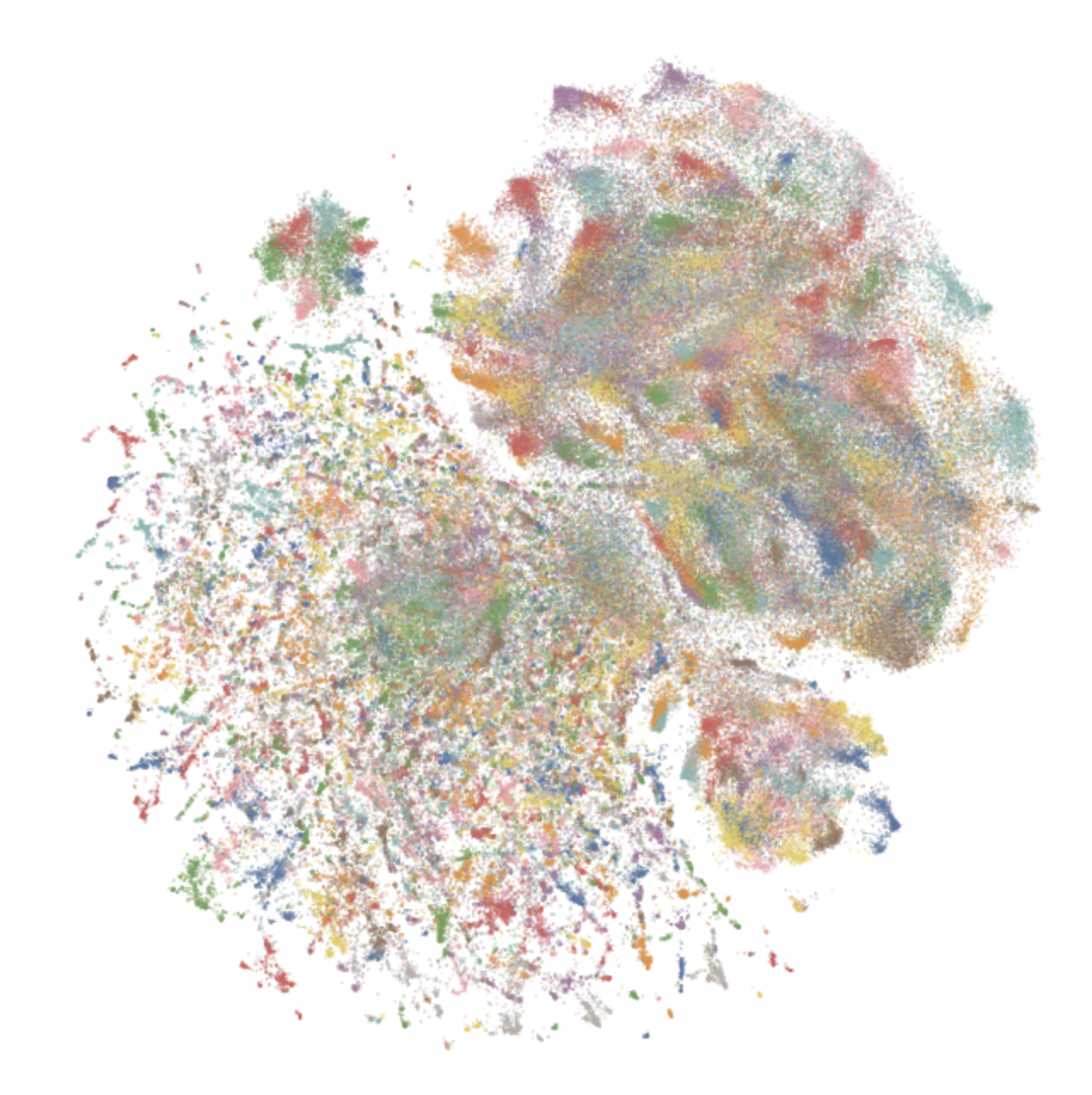}}
        \caption{}
        \label{subfig:gpt4all-data}
    \end{subfigure}
    \begin{subfigure}[b]{.55\columnwidth}
        \centering
        \href{https://atlas.nomic.ai/map/gpt4all-j-prompts-curated}{\includegraphics[width=1\textwidth]{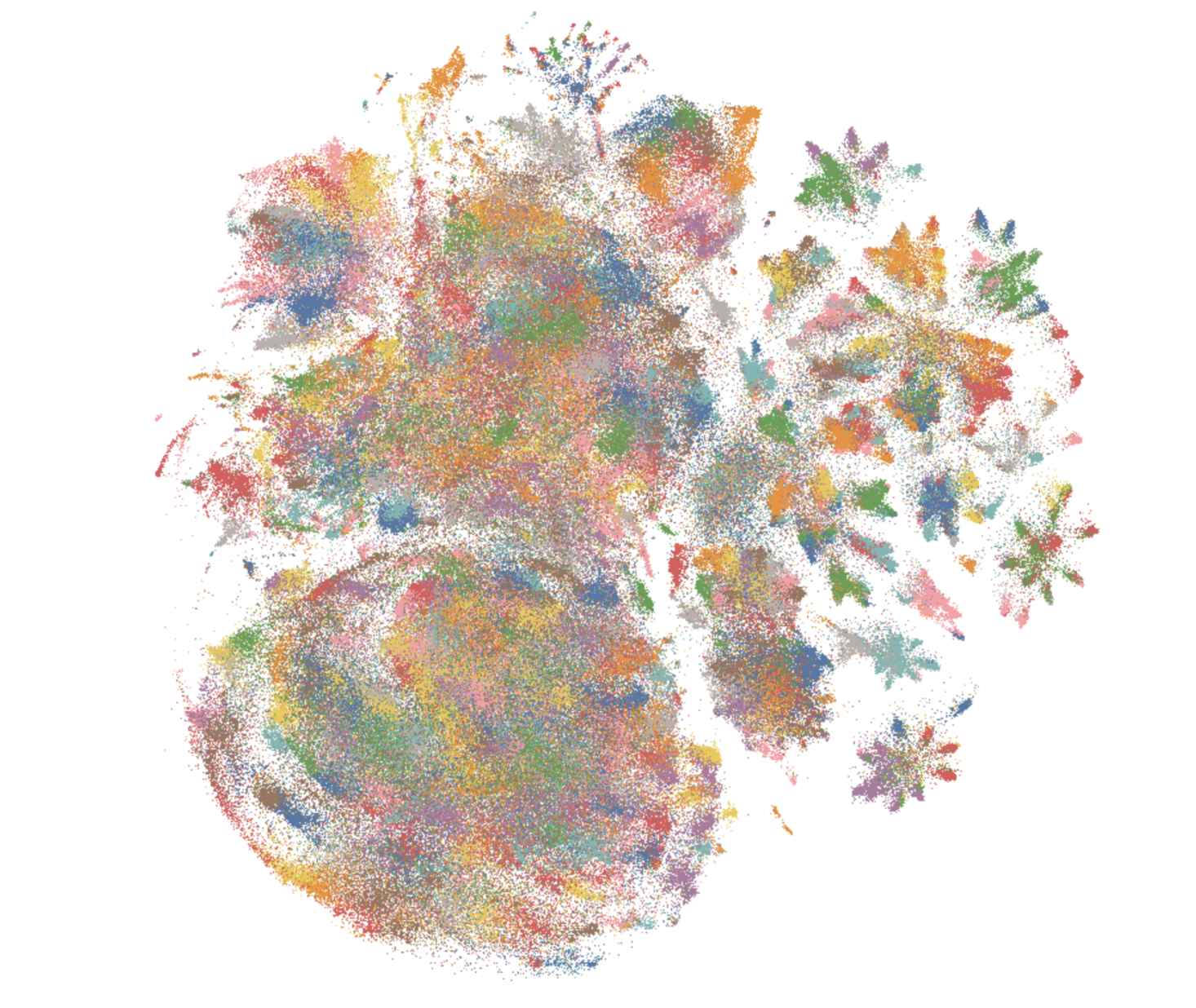}}
        \caption{}
        \label{fig:gpt4all-}
    \end{subfigure}
    \begin{subfigure}[b]{.5\columnwidth}
        \centering
        \href{https://atlas.nomic.ai/map/gpt4all-deduped-prompt}{\includegraphics[width=1\textwidth]{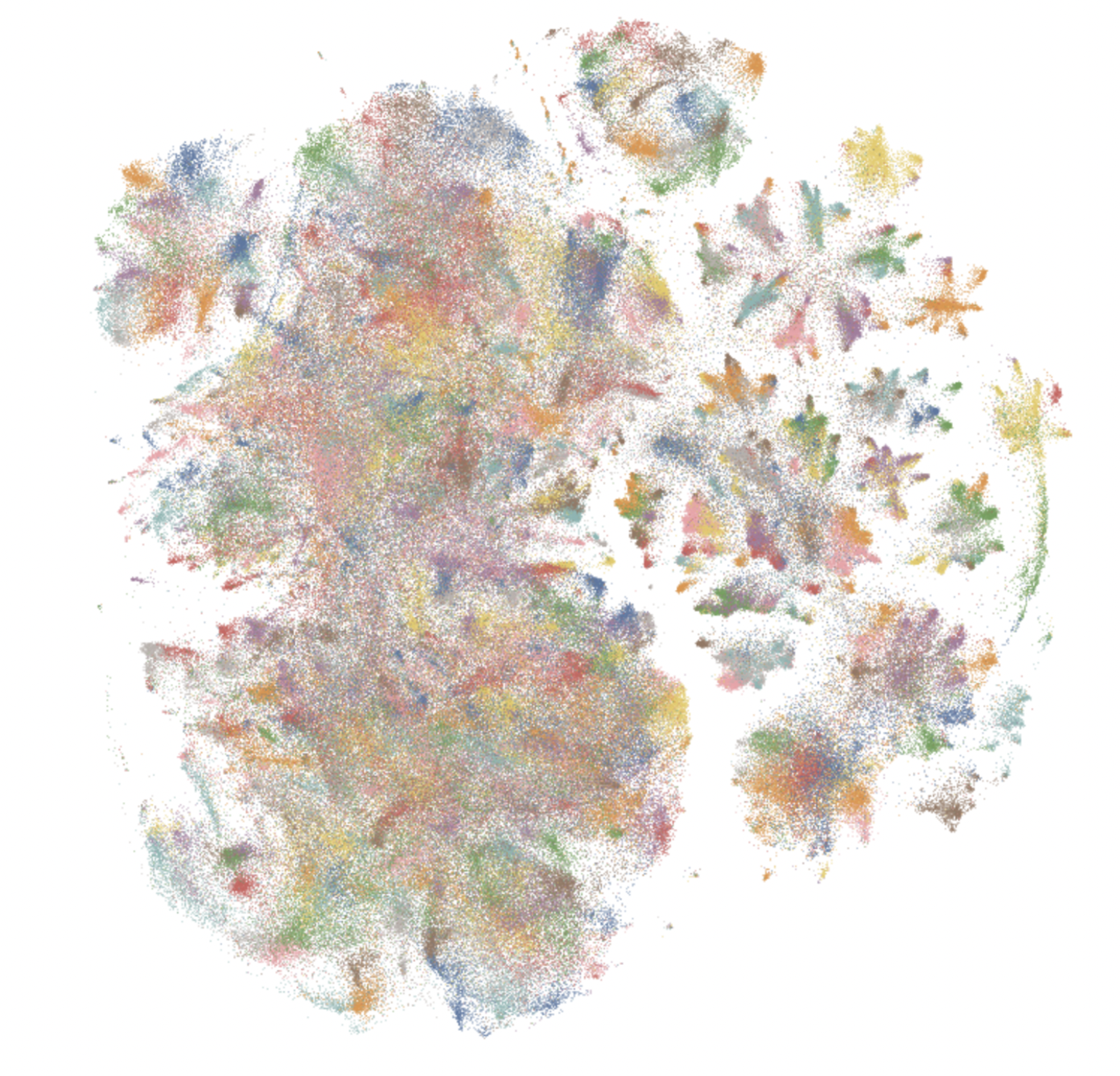}}
        \caption{}
        \label{fig:gpt4all-snoozy}
    \end{subfigure}
    \label{fig:train_data}
    \caption{TSNE visualizations showing the progression of the GPT4All train set. Panel (a) shows the original uncurated data. The red arrow denotes a region of highly homogeneous prompt-response pairs. The coloring denotes which open dataset contributed the prompt. Panel (b) shows the original GPT4All data after curation. This panel, as well as panels (c) and (d) are 10 colored by topic, which Atlas automatically extracts. Notice that the large homogeneous prompt-response blobs no longer appearl. Panel (c) shows the GPT4All-J dataset. The "starburst" clusters introduced on the right side of the panel correspond to the newly added creative data. Panel (d) shows the final GPT4All-snoozy dataset. 
    All datasets have been released to the public, and can be interactively explored online.
    In the web version of this article, you can click on a panel to be taken to its interactive visualization.
    }
\end{figure*}

The collected dataset was loaded into Atlas~\cite{atlas-nomic-ai}---a visual interface for exploring and tagging massive unstructured datasets ---for data curation.
Using Atlas, we identified and removed subsets of the data where GPT-3.5-Turbo refused to respond, had malformed output, or produced a very short response.
This resulted in the removal of the entire Bigscience/P3 subset of our data, as many P3 prompts induced responses that were simply one word. 
After curation, we were left with a set of 437,605 prompt-response pairs, which we visualize in Figure~\ref{subfig:dirty}.

\subsection{Model Training}
The original GPT4All model was a fine tuned variant of LLaMA 7B.
In order to train it more efficiently, we froze the base weights of LLaMA, and only trained a small set of LoRA~\cite{hu2021lora} weights during the fine tuning process.
Detailed model hyper-parameters and training code can be found in our associated code repository\footnote{\href{https://github.com/nomic-ai/gpt4all}{https://github.com/nomic-ai/gpt4all}}.

\subsection{Model Access}
We publicly released all data, training code, and model weights for the community to build upon.
Further, we provided a 4-bit quantized version of the model, which enabled users to run it on their own commodity hardware without transferring data to a 3rd party service.

Our research and development costs were dominated by $\sim$\$800 in GPU spend (rented from Lambda Labs and Paperspace) and $\sim$\$500 in OpenAI API spend.
Our final GPT4All model could be trained in about eight hours on a Lambda Labs DGX A100 8x 80GB for a total cost of $\sim$\$100.

\subsection{Model Evaluation}
We performed a preliminary evaluation of our model using the human evaluation data from the Self Instruct paper~\cite{wang2023selfinstruct}.
We reported the ground truth perplexity of our model against what was, to our knowledge, the best openly available alpaca-lora model at the time, provided by user \textit{chainyo} on HuggingFace.
Both models had very large perplexities on a small number of tasks, so we reported perplexities clipped to a maximum of 100.
We found that GPT4All produces stochastically lower ground truth perplexities than alpaca-lora~\cite{gpt4all}.

\begin{table*}[!h]
\fontsize{8.5pt}{8.5pt}\selectfont
\centering
\begin{tabular}{lcccccccc}
\toprule
Model & BoolQ & PIQA & HellaSwag & WinoG. & ARC-e & ARC-c & OBQA & Avg. \\
\midrule
GPT4All-J 6B v1.0* & 73.4 & 74.8 & 63.4 & 64.7 & 54.9 & 36 & 40.2 & 58.2 \\
GPT4All-J v1.1-breezy* & 74 & 75.1 & 63.2 & 63.6 & 55.4 & 34.9 & 38.4 & 57.8 \\
GPT4All-J v1.2-jazzy* & 74.8 & 74.9 & 63.6 & 63.8 & 56.6 & 35.3 & 41 & 58.6 \\
GPT4All-J v1.3-groovy* & 73.6 & 74.3 & 63.8 & 63.5 & 57.7 & 35 & 38.8 & 58.1 \\
GPT4All-J Lora 6B* & 68.6 & 75.8 & 66.2 & 63.5 & 56.4 & 35.7 & 40.2 & 58.1 \\
GPT4All LLaMa Lora 7B* & 73.1 & 77.6 & 72.1 & 67.8 & 51.1 & 40.4 & 40.2 & 60.3 \\
GPT4All 13B snoozy* & 83.3 & 79.2 & 75 & 71.3 & 60.9 & 44.2 & 43.4 & 65.3 \\
GPT4All Falcon & 77.6 & 79.8 & 74.9 & 70.1 & 67.9 & 43.4 & 42.6 & 65.2 \\
Nous-Hermes~\cite{nousresearch2023noushermes} & 79.5 & 78.9 & 80 & 71.9 & 74.2 & 50.9 & \textbf{46.4} & 68.8 \\
Nous-Hermes2~\cite{nousresearch2023noushermesllama} & \textbf{83.9} & \textbf{80.7} & 80.1 & 71.3 & 75.7 & \textbf{52.1} & 46.2 & \textbf{70.0} \\
Nous-Puffin~\cite{nousresearch2023redmondpuffin} & 81.5 & \textbf{80.7} & \textbf{80.4} & \textbf{72.5} & \textbf{77.6} & 50.7 & 45.6 & 69.9 \\
Dolly 6B*~\cite{DatabricksBlog2023DollyV1} & 68.8 & 77.3 & 67.6 & 63.9 & 62.9 & 38.7 & 41.2 & 60.1 \\
Dolly 12B*~\cite{DatabricksBlog2023DollyV2} & 56.7 & 75.4 & 71 & 62.2 & 64.6 & 38.5 & 40.4 & 58.4 \\
Alpaca 7B*~\cite{alpaca} & 73.9 & 77.2 & 73.9 & 66.1 & 59.8 & 43.3 & 43.4 & 62.5 \\
Alpaca Lora 7B*~\cite{alpaca-lora} & 74.3 & 79.3 & 74 & 68.8 & 56.6 & 43.9 & 42.6 & 62.8 \\
GPT-J* 6.7B~\cite{gpt-j}  & 65.4 & 76.2 & 66.2 & 64.1 & 62.2 & 36.6 & 38.2 & 58.4 \\
LLama 7B*~\cite{touvron2023llama} & 73.1 & 77.4 & 73 & 66.9 & 52.5 & 41.4 & 42.4 & 61.0 \\
LLama 13B*~\cite{touvron2023llama} & 68.5 & 79.1 & 76.2 & 70.1 & 60 & 44.6 & 42.2 & 63.0 \\
Pythia 6.7B*~\cite{biderman2023pythia}  & 63.5 & 76.3 & 64 & 61.1 & 61.3 & 35.2 & 37.2 & 56.9 \\
Pythia 12B*~\cite{biderman2023pythia} & 67.7 & 76.6 & 67.3 & 63.8 & 63.9 & 34.8 & 38 & 58.9 \\
Fastchat T5*~\cite{zheng2023judging} & 81.5 & 64.6 & 46.3 & 61.8 & 49.3 & 33.3 & 39.4 & 53.7 \\
Fastchat Vicuña* 7B~\cite{zheng2023judging} & 76.6 & 77.2 & 70.7 & 67.3 & 53.5 & 41.2 & 40.8 & 61.0 \\
Fastchat Vicuña 13B*~\cite{zheng2023judging} & 81.5 & 76.8 & 73.3 & 66.7 & 57.4 & 42.7 & 43.6 & 63.1 \\
StableVicuña RLHF*~\cite{stabilityai2023stablelm} & 82.3 & 78.6 & 74.1 & 70.9 & 61 & 43.5 & 44.4 & 65.0 \\
StableLM Tuned*~\cite{stabilityai2023stablelm} & 62.5 & 71.2 & 53.6 & 54.8 & 52.4 & 31.1 & 33.4 & 51.3 \\
StableLM Base*~\cite{stabilityai2023stablelm} & 60.1 & 67.4 & 41.2 & 50.1 & 44.9 & 27 & 32 & 46.1 \\
Koala 13B*~\cite{koala_blogpost_2023} & 76.5 & 77.9 & 72.6 & 68.8 & 54.3 & 41  & 42.8 & 62.0 \\
Open Assistant Pythia 12B* & 67.9 & 78   & 68.1 & 65   & 64.2 & 40.4 & 43.2 & 61.0 \\
Mosaic MPT7B~\cite{MosaicML2023Introducing} & 74.8 & 79.3 & 76.3 & 68.6 & 70   & 42.2 & 42.6 & 64.8 \\
Mosaic mpt-instruct~\cite{MosaicML2023Introducing}  & 74.3 & 80.4 & 77.2 & 67.8 & 72.2 & 44.6 & 43   & 65.6 \\
Mosaic mpt-chat~\cite{MosaicML2023Introducing} & 77.1 & 78.2 & 74.5 & 67.5 & 69.4 & 43.3 & 44.2 & 64.9 \\
Wizard 7B~\cite{xu2023wizardlm} & 78.4 & 77.2 & 69.9 & 66.5 & 56.8 & 40.5 & 42.6 & 61.7 \\
Wizard 7B Uncensored~\cite{xu2023wizardlm} & 77.7 & 74.2 & 68   & 65.2 & 53.5 & 38.7 & 41.6 & 59.8 \\
Wizard 13B Uncensored~\cite{xu2023wizardlm} & 78.4 & 75.5 & 72.1 & 69.5 & 57.5 & 40.4 & 44   & 62.5 \\
GPT4-x-Vicuna-13b~\cite{nousresearch2023gpt4xvicuna} & 81.3 & 75   & 75.2 & 65   & 58.7 & 43.9 & 43.6 & 63.2 \\
Falcon 7b~\cite{falcon40b} & 73.6 & \textbf{80.7} & 76.3 & 67.3 & 71   & 43.3 & 44.4 & 65.2 \\
Falcon 7b instruct~\cite{falcon40b} & 70.9 & 78.6 & 69.8 & 66.7 & 67.9 & 42.7 & 41.2 & 62.5 \\
\midrule
text-davinci-003 & 88.1 & 83.8 & 83.4 & 75.8 & 83.9 & 63.9 & 51.0 & 75.7 \\
\bottomrule
\end{tabular}
\caption{Evaluations of all language models in the GPT4All ecosystem as of August 1, 2023. Code models are not included. OpenAI's text-davinci-003 is included as a point of comparison. The best overall performing model in the GPT4All ecosystem, Nous-Hermes2, achieves over 92\% of the average performance of text-davinci-003. Models marked with an asterisk were available in the ecosystem as of the release of GPT4All-Snoozy. Note that at release, GPT4All-Snoozy had the best average performance of any model in the ecosystem. Bolded numbers indicate the best performing model as of August 1, 2023. }
\label{table:evals}
\end{table*}

\section{From a Model to an Ecosystem}
\subsection{GPT4All-J: Repository Growth and the implications of the LLaMA License}
\begin{figure*}[t]
     \centering
     \includegraphics[width=0.60\linewidth]{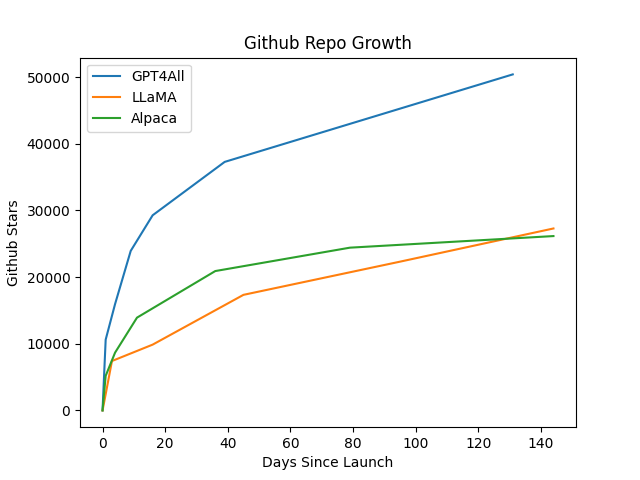}
     \caption{Comparison of the github start growth of GPT4All, Meta's LLaMA, and Stanford's Alpaca. We conjecture that GPT4All achieved and maintains faster ecosystem growth due to the focus on access, which allows more users to meaningfully participate.}
     \label{fig:star-chart}
 \end{figure*}

The GPT4All repository grew rapidly after its release, gaining over 20000 GitHub stars in just one week, as shown in Figure~\ref{fig:star-chart}.
This growth was supported by an in-person hackathon hosted in New York City three days after the model release, which attracted several hundred participants.
As the Nomic discord, the home of online discussion about GPT4All, ballooned to over 10000 people, one thing became very clear - there was massive demand for a model that could be used commercially.

The LLaMA model that GPT4All was based on was licensed for research only, which severely limited the set of domains that GPT4All could be applied in.
As a response to this, the Nomic team repeated the model training procedure of the original GPT4All model, but based on the already open source and commercially licensed GPT-J model~\cite{gpt-j}.
GPT4All-J also had an augmented training set, which contained multi-turn QA examples and creative writing such as poetry, rap, and short stories.
The creative writing prompts were generated by filling in schemas such as "Write a [CREATIVE STORY TYPE] about [NOUN] in the style of [PERSON]."
We again employed Atlas to curate the prompt-response pairs in this data set.

Our evaluation methodology also evolved as the project grew.
In particular, we began evaluating GPT4All models using a suite of seven reasoning tasks that were used for evaluation of the Databricks Dolly~\cite{DatabricksBlog2023DollyV2} model, which was released on April 12, 2023.
Unfortunately, GPT4All-J did not outperform other prominent open source models on this evaluation.
As a result, we endeavoured to create a model that did.

\subsection{GPT4All-Snoozy: the Emergence of the GPT4All Ecosystem}
GPT4All-Snoozy was developed using roughly the same procedure as the previous GPT4All models, but with a few key modifications.
First, GPT4All-Snoozy used the LLaMA-13B base model due to its superior base metrics when compared to GPT-J.
Next, GPT4All-Snoozy incorporated the Dolly's training data into its train mix.
After data curation and deduplication with Atlas, this yielded a training set of 739,259 total prompt-response pairs.
We dubbed the model that resulted from training on this improved dataset GPT4All-Snoozy.
As shown in Figure \ref{table:evals}, GPT4All-Snoozy had the best average score on our evaluation benchmark of any model in the ecosystem at the time of its release.

Concurrently with the development of GPT4All, several organizations such as LMSys, Stability AI, BAIR, and Databricks built and deployed open source language models.
We heard increasingly from the community that they wanted quantized versions of these models for local use.
As we realized that organizations with ever more resources were developing source language models, we decided to pivot our effort away from training increasingly capable models and towards providing easy access to the plethora of models being produced by the open source community.
Practically, this meant spending our time compressing open source models for use on commodity hardware, providing stable and simple high level model APIs, and supporting a GUI for no code model experimentation.

\subsection{The Current State of GPT4All}

Today, GPT4All is focused on improving the accessibility of open source language models.
The repository provides compressed versions of open source models for use on commodity hardware, stable and simple high level model APIs, and a GUI for no code model experimentation.
The project continues to increase in popularity, and as of August 1 2023, has garnered over 50000 GitHub stars and over 5000 forks.

GPT4All currently provides native support and benchmark data for over 35 models (see Figure~\ref{table:evals}), and includes several models co-developed with industry partners such as Replit and Hugging Face.
GPT4All also provides high level model APIs in languages including Python, Typescript, Go, C\#, and Java, among others.
Furthermore, the GPT4All no code GUI currently supports the workflows of over 50000 monthly active users, with over 25\% of users coming back to the tool every day of the week.
(Note that all GPT4All user data is collected on an \textit{opt in} basis.)
GPT4All has become the top language model integration in the popular open source AI orchestration library LangChain~\cite{langchain}, and powers many popular open source projects such as PrivateGPT~\cite{privategpt}, Quiver~\cite{stangirard2023quivr}, and MindsDB~\cite{mindsdb}, among others.
GPT4All is the 3rd fastest growing GitHub repository of all time~\cite{leo_github_fastest}, and is the 185th most popular repository on the platform, by star count.

\section{The Future of GPT4All}
In the future, we will continue to grow GPT4All, supporting it as the de facto solution for LLM accessibility.
Concretely, this means continuing to compress and distribute important open-source language models developed by the community, as well as compressing and distributing increasingly multimodal AI models.
Furthermore, we will expand the set of hardware devices that GPT4All models run on, so that GPT4All models ``just work" on any machine, whether it comes equipped with Apple Metal silicon, NVIDIA, AMD, or other edge-accelerated hardware.
Overall, we envision a world where anyone, anywhere, with any machine, can access and contribute to the cutting edge of AI.

\section*{Limitations}
By enabling access to large language models, the GPT4All project also inherits many of the ethical concerns associated with generative models.
Principal among these is the concern that unfiltered language models like GPT4All enable malicious users to generate content that could be harmful and dangerous (e.g., instructions on building bioweapons).
While we recognize this risk, we also acknowledge the risk of concentrating this technology in the hands of a limited number of increasingly secretive research groups.
We believe that the risk of focusing on the benefits of language model technology significantly outweighs the risk of misuse, and hence we prefer to make the technology as widely available as possible.

Finally, we realize the challenge in assigning credit for large-scale open source initiatives.
We make a first attempt at fair credit assignment by explicitly including the GPT4All open source developers as authors on this work, but recognize that this is insufficient fully characterize everyone involved in the GPT4All effort.
Furthermore, we acknowledge the difficulty in citing open source works that do not necessarily have standardized citations, and do our best in this paper to provide URLs to projects whenever possible.
We encourage further research in the area of open source credit assignment, and hope to be able to support some of this research ourselves in the future.

\bibliography{emnlp2023}
\bibliographystyle{acl_natbib}

\end{document}